\documentclass[journal]{IEEEtran}

\usepackage{xcolor,soul,framed} 

\colorlet{shadecolor}{yellow}
\usepackage[pdftex]{graphicx}
\graphicspath{{../pdf/}{../jpeg/}{./img/}}
\DeclareGraphicsExtensions{.pdf,.jpeg,.png,.jpg}

\usepackage[cmex10]{amsmath}
\usepackage{array}
\usepackage{mdwmath}
\usepackage{mdwtab}
\usepackage{eqparbox}
\usepackage{url}
\usepackage{graphicx}
\usepackage{amsmath}
\usepackage{algorithm}
\usepackage{algorithmic}
\usepackage{multirow}
\usepackage{booktabs}
\usepackage{wrapfig}
\usepackage{subcaption}

\begin{document}
\bstctlcite{IEEEexample:BSTcontrol}
    \title{ES-Net: Erasing Salient Parts to Learn More in Re-Identification}
  \author{
      Dong Shen, Shuai Zhao, Jinming Hu, Hao Feng, Deng Cai~\IEEEmembership{Member,~IEEE,}\\
      Xiaofei He~\IEEEmembership{Senior Member,~IEEE,}
	  \thanks{ This work was supported in part by the National Key Research and Development Program of China under Grant 2018AAA0101400; in part by the National Nature Science Foundation of China under Grant 62036009, Grant U1909203, and Grant 61936006; and in part by the Alibaba-Zhejiang University Joint Institute of Frontier Technologies.(Corresponding author:Deng Cai.)}
	  \thanks{Dong Shen, Shuai Zhao, Jinming Hu, Hao Feng, and Xiaofei He are with the State Key Laboratory of Computer-Aided Design (CAD) and Computer Graphics (CG), Zhejiang University, Hangzhou 310027, China (e-mail: seanshen@zju.edu.cn).}
	  \thanks{Deng Cai is with the State Key Laboratory of Computer-Aided Design (CAD) and Computer Graphics (CG), Zhejiang University, Hangzhou 310027, China, and also with the Alibaba-Zhejiang University Joint Institute of Frontier Technologies, 866 Yuhangtang Road, Xihu District, Hangzhou 310027, China. (e-mail: dengcai@cad.zju.edu.cn).}
  }



\maketitle

\begin{abstract}
As an instance-level recognition problem, re-identification (re-ID) requires models to capture diverse features. However, with continuous training, re-ID models pay more and more attention to the salient areas. As a result, the model may only focus on few small regions with salient representations and ignore other important information. This phenomenon leads to inferior performance, especially when models are evaluated on small inter-identity variation data. In this paper, we propose a novel network, Erasing-Salient Net (ES-Net), to learn comprehensive features by erasing the salient areas in an image. ES-Net proposes a novel method to locate the salient areas by the confidence of objects and erases them efficiently in a training batch. Meanwhile, to mitigate the over-erasing problem, this paper uses a trainable pooling layer P-pooling that generalizes global max and global average pooling. Experiments are conducted on two specific re-identification tasks (i.e., Person re-ID, Vehicle re-ID). Our ES-Net outperforms state-of-the-art methods on three Person re-ID benchmarks and two Vehicle re-ID benchmarks. Specifically, mAP / Rank-1 rate: 88.6\% / 95.7\% on Market1501, 78.8\% / 89.2\% on DuckMTMC-reID,  57.3\% / 80.9\% on MSMT17, 81.9\% / 97.0\% on Veri-776, respectively. Rank-1 / Rank-5 rate: 83.6\% / 96.9\% on VehicleID (Small), 79.9\% / 93.5\% on VehicleID (Medium),  76.9\% / 90.7\% on VehicleID (Large), respectively. Moreover, the visualized salient areas show human-interpretable visual explanations for the ranking results. 
\end{abstract}


%
\IEEEpeerreviewmaketitle


\section{Introduction}

\IEEEPARstart{P}{erson} 
re-identification and vehicle re-identification aim to retrieve the images of the same ID. It has attracted much attention recently due to its wide applications in surveillance. Though tremendous progress has been achieved in the last few years \cite{sun2018beyond,yang2019towards,ye2020cross}, re-ID is still a challenging task because of many practical problems, e.g., occlusion, pose variation, viewpoint changes, and background clutters.

\begin{figure}[t]
    \centering
    \includegraphics[width=\linewidth]{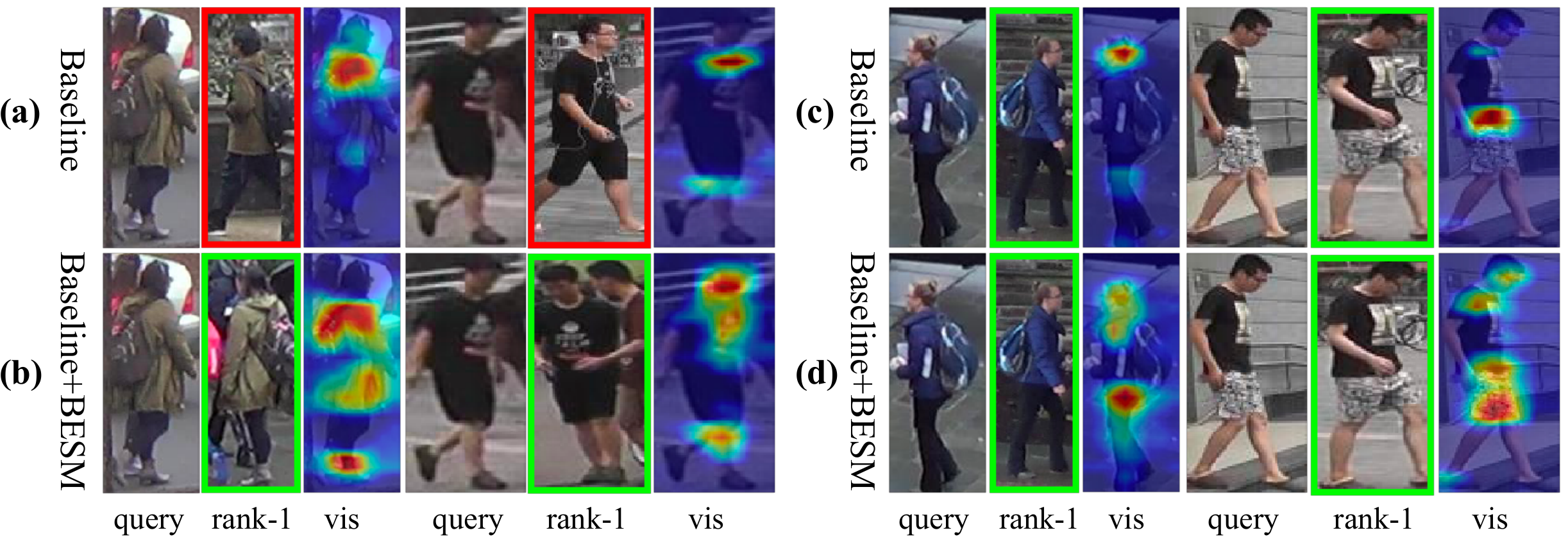}
    \caption{Learning comprehensive features benefits query accuracy. The "vis" is the visualization results of the salient areas in the query images, which are found by proposed CG-RAM. BESM means our erasing method, named Batch Erasing Salient Method. Images with green and red boundary denote true positive and false positive. (a) and (b) show that our method helps model find the correct rank-1 by discovering more comprehensive features. In (c) and (d), our method and baseline give the same rank-1 results for the same query images. However, our method discovers more diverse salient areas, which further demonstrates that our method does not succeed by accident but definitely learns more comprehensive features than the baseline.
    }
    \vspace{-0.4cm}
    \label{fig:viscmp}
    \centering
    
\end{figure}

During continuous training, re-ID models tend to pay more and more attention to the most salient areas. As a result, the trained model may only focus on few small regions with salient representations and neglect other important information, which may be enough to distinguish different identities with large visual differences. However, it is common that the model fails to find the right images with slight inter-identity differences during evaluation. As illustrated in Figure \ref{fig:viscmp}a, the baseline only focuses on the few small salient areas and gives the wrong rank-1 results given the query images. In the first pair in Figure \ref{fig:viscmp}a, the baseline focuses on $green\;clothes$ but neglects other discriminative visual features, e.g., $shoes$. As can be seen, it is essential for the re-ID model to capture diverse information to learn comprehensive features.

 
Recent studies tried to learn comprehensive features can be mainly grouped into three categories: 1) some methods \cite{dai2019batch,zhong2017random} utilize the erasing method by randomly erases a part of the feature maps of input images to force the model to learn diverse features; 2) some part-based works \cite{sun2018beyond,wang2018learning} assemble the salient features of different human parts and global features to form a more comprehensive descriptor; 3) some \cite{yang2019towards,si2018dual,chen2020salience} use the attention mechanism to discover important and comprehensive features.

\begin{figure*}[t]
	\centering
		\includegraphics[width=1.9\columnwidth]{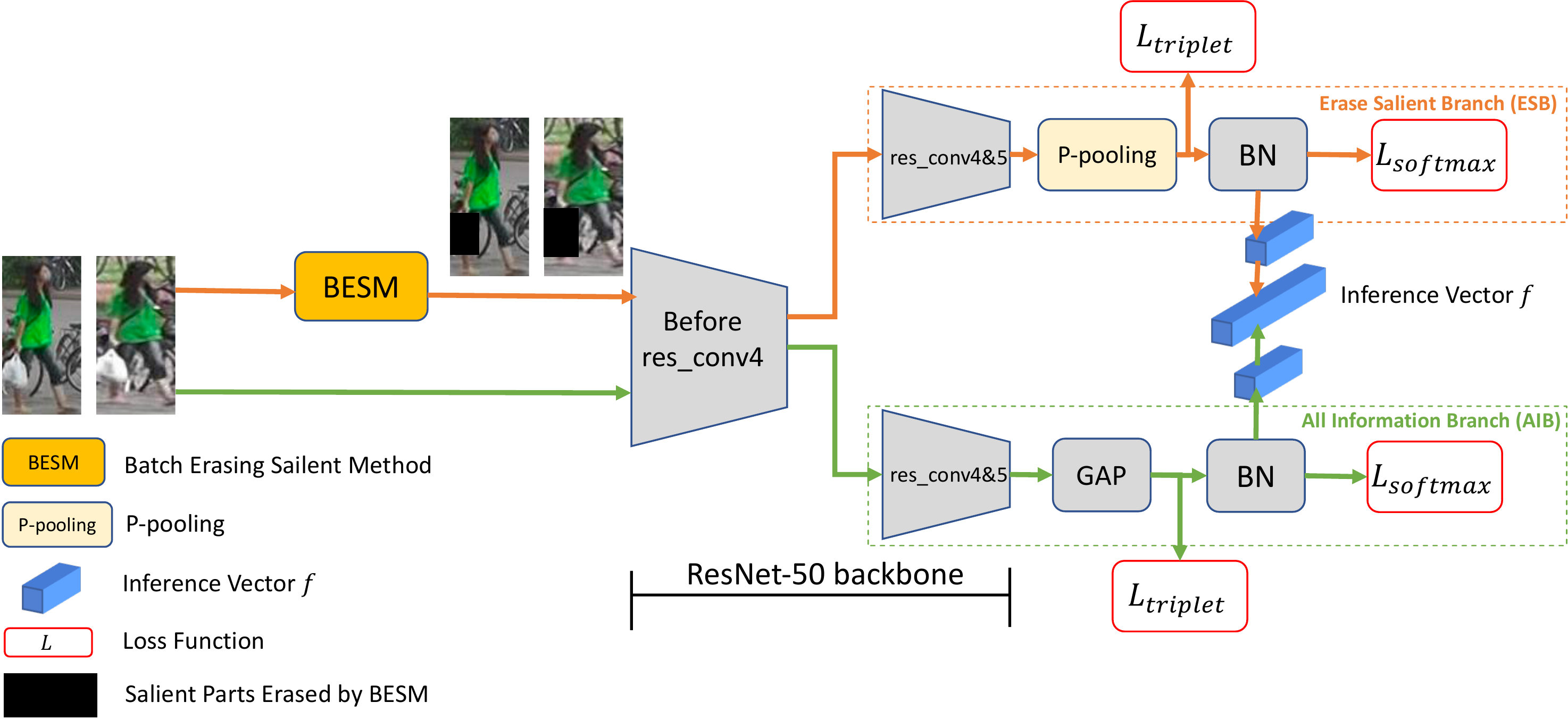}
		\caption{Overview of our ES-Net. ES-Net consists of two branches: AIB and ESB. ESB utilizes BESM to make the model learn more by erasing the salient parts and use P-pooling to mitigate the over-erasing problem. AIB is trained by the entire images to keep important information. 
		}
		\label{figoverview}
	\end{figure*}

The proposed ES-Net utilizes the erasing mechanism; it erases the salient regions of the images to force models to learn more. Compared with the randomly erasing method \cite{zhong2017random,dai2019batch}, our method provides interpretable visualization results and clear erasing guidance by utilizing the salient maps. ES-Net consists of two branches: All Information Branch (AIB) and Erasing Salient Branch. ESB learns comprehensive information by erasing the salient regions of the input images. AIB utilizes the entire images to prevent losing the discriminative information. The overall architecture is shown in Figure \ref{figoverview}.

To erase the salient regions, we have to know which regions play an essential role in the retrieval process. However, most of the existing methods of locating the salient regions are designed for the classification task, e.g., CAM \cite{zhou2016learning}, Grad-CAM \cite{selvaraju2017grad}, and Grad-CAM++ \cite{chattopadhay2018grad}. As for re-ID, commonly used methods RAM \cite{yang2019towards} and \cite{zheng2019re} can only be applied to specific model structures or a certain layer. Hence, this paper proposes a novel method, termed Confidence Gradient-weighted Ranking Activation Mapping (CG-RAM), to locate the salient region for re-ID. CG-RAM utilizes the gradients of the confidence score, which represents whether the query and galley images belong to the same identity, with respect to the feature maps to get the salient maps. Our CG-RAM can be applied to any layer of any model structure and shows human-interpretable visual explanations for the ranking results. Based on CG-RAM, an erasing method, termed Batch Erasing Salient Method (BESM), is proposed. BESM forces the model to learn more by erasing the salient regions of the inputs images efficiently in mini-batch training.

Meanwhile, the model may not be able to extract the robust features to represent the identity if too many salient areas are erased, especially when Global Average Pooling (GAP) is applied in the model. GAP is a commonly used operation in re-ID models; it takes the feature map as input and outputs the mean value of each channel. After erasing the salient parts, GAP may lose valuable information because the remaining areas are not distinguishable as before. By contrast, Global Max Pooling (GMP) calculates the max value, which enables GMP to find the salient areas effectively, but GMP is sensitive to noise data. As can be seen, both GAP and GMP are not suitable when the over-erasing problem occurs in training. To mitigate the over-erasing problem, we use P-pooling \cite{Wei_2019_CVPR}, which is designed for segmentation tasks. P-pooling generalizes GAP and GMP, which makes P-pooling have good robustness like GAP and enables it to extract salient information like GMP.

Our contributions can be summarized as the following:
\begin{itemize}
  \item A novel network ES-Net is proposed to discover diverse features by erasing salient areas. Extensive experimental results on five datasets of two specific re-identification tasks (i.e., Person re-ID, Vehicle re-ID) show that ES-Net achieves better performance than state-of-the-art methods.
  \item  A new method CG-RAM is proposed to find the salient areas for the re-ID task. CG-RAM can be applied to any layer of any model structure and can also be used to visualize and interpret the results in the test stage. 
  \item Based on CG-RAM, BESM is proposed to erase the salient areas of the images efficiently in training. Besides, we use P-pooling to alleviate the over-erasing problem.
\end{itemize}

\section{Related work}
\textbf{Person re-ID} 
Considering different tasks, we divide most of the re-ID methods into three categories, i.e., image-based, video-based and multi-modality. Video-based methods\cite{hou2019vrstc,ye2019dynamic} diverse person samples in consecutive frames. Multi-modality methods\cite{wu2017robust,ye2020cross} combine the multiple modality information to improve the performance. In this paper, we focus on the image-based re-ID.

Person re-ID consists of 2 steps: obtaining a feature embedding and performing matching. This paper focuses on the former. It is essential for getting a high-quality and robust feature embedding to capture diverse features. Lots of works have been proposed to help model find comprehensive features. Some part-based methods \cite{sun2018beyond,wang2018learning} assemble the salient features from different parts and the global feature to form a more comprehensive descriptor, e.g., MGN\cite{wang2018learning} proposes an end-to-end feature learning strategy integrating discriminative information with various granularities. Some utilize the erasing mechanism, e.g., Random Erasing \cite{zhong2017random} drops parts of the input images randomly; BDB \cite{dai2019batch} randomly drops the same region of all input feature maps in a batch to reinforce the attentive feature learning of local regions. 
Some works\cite{zheng2019joint,ye2020augmentation} improve learned embeddings by data augmentation.
Some papers \cite{si2018dual,li2018harmonious} utilize attention mechanisms to discover salient and comprehensive features, e.g., CAMA \cite{yang2019towards} uses attention-based loss to punish the overlapped parts; DuATM\cite{si2018dual} uses a dual attention mechanism to learn context-aware feature sequences and perform attentive sequence comparison simultaneously.

\textbf{Vehicle re-ID} 
Vehicle re-identification requires robust and discriminative image representation. Some approaches use extra attribute information, e.g., viewpoint, color to guide model training. For instance, VA-Net \cite{chu2019vehicle} proposes a viewpoint-aware metric learning approach that learns two metrics for similar viewpoints and different viewpoints in two feature spaces. Yan \emph{et al.} \cite{yan2017exploiting} proposes two ranking methods, generalized pairwise ranking and multigrain based list ranking based on the multigrain constraints. Some works utilize the temporal-spatial information during the training and testing stage, e.g., Shen \emph{et al.} \cite{shen2017learning} use Siamese-CNN and Path-LSTM to incorporates important visual-spatial-temporal path information for regularization.

\textbf{Locating Salient Areas}
An expected salient areas locating method should reveal which pixels are more critical in the image. It can effectively help us analyze the model and improve the results. However, most of the existing methods are designed for the classification task, e.g., CAM \cite{zhou2016learning}, Grad-CAM \cite{selvaraju2017grad}, and Grad-CAM++ \cite{chattopadhay2018grad}. Grad-CAM and Gram-CAM++ use the gradients of the target label to produce a localization map highlighting the important regions in the image under the supervision of classification loss. As for re-ID, there are mainly two methods. One is called RAM \cite{yang2019towards}, and it requires a GAP between the final descriptor and the feature maps where we locate the salient areas. The other method \cite{zheng2019re} uses the siamese network to convert the retrieval into the classification task, so it can only be used in the siamese network.


\textbf{Global Pooling Operation}
Global pooling is an essential operation in re-ID; it takes the feature maps as input and outputs the vectors, which directly influences the global image descriptors. GAP \cite{babenko2015aggregating} calculates the mean value, and it is widely used in the re-ID models. GMP \cite{razavian2016visual} outputs the max value of each channel, which enables GMP to extract the salient information effectively, but it is sensitive to the noisy data. R-MAC \cite{DBLP:journals/corr/ToliasSJ15} method calculates a linear combination of max and sum pooling. P-pooling \cite{Wei_2019_CVPR} finds an intermediate form between max and average pooling to provide a balanced and self-adjusted pooling strategy for segmentation.


\section{Approach}

In this section, we first give an overview of the proposed ES-Net. The overview architecture of ES-Net is shown in Figure \ref{figoverview}. Then we introduce CG-RAM and BESM to locate and erase the salient parts. In the end, we describe P-pooling, which is used to mitigate the over-erasing problem.
\subsection{Network Overview}
The key idea of ES-Net is to capture diverse features by erasing the salient parts. It consists of two branches: AIB and ESB. AIB is trained with the entire images to keep the salient features. ESB utilizes BESM to make model learn more. To avoid ESB collapsing because of the over-erasing problem, we use P-pooling instead of GAP in ESB. We choose ResNet-50 \cite{he2016deep} as the backbone and use the techniques discussed in the \cite{luo2019bag}: setting the stride of the last residual block to 1 and adding the BN layer. ESB shares the weight of layers before the $res\_conv4$ with AIB. We use the cross-entropy loss and the hard mining triplet loss to form final loss function $L_{total}$.
\begin{align}
\label{loss}
  L_{total} =& L_{ESB}^{id} + L_{ESB}^{triplet} + L_{AIB}^{id} + L_{AIB}^{triplet}.
\end{align}%

\subsection{Gradient-weighted Ranking Activation Mapping}

To erase the salient parts, we must find them first. Inspired by Grad-CAM, we propose the CG-RAM to find the salient regions for re-ID. CG-RAM can be applied in any layer and has no requirements for the model structure.

Let $q$ and $g$ denote the query and the gallery image, respectively. We also denote the query vector and the gallery vector, which are used to retrieve during testing, by $f_q$ and $f_g$, respectively. $S_{q,g}$ indicates the confidence score of whether the query picture and the galley picture belong to the same identity. Here we use cosine distance of $f_q$ and $f_g$ to represent $S_{q,g}$,
\begin{align}
\label{score}
S_{q,g} =& \frac{\left< f_{q}, f_{g}\right>}{\left|\left| f_{q}\right|\right|\left|\left| f_{g}\right|\right|}.
\end{align}%

To get the salient map $M$, we first compute the gradient $ \phi_{i j}^k$ of the score $S_{q,g}$ with respect to the feature value $A^{k}_{i j}$, $A$ is the feature maps of a convolutional layer where we want to find the salient regions, $k$ means the $k^{th}$ channel of the feature maps, $(i,j)$ denotes the spatial location in the feature maps.
\begin{align}
\label{weight}
    \phi_{i j}^k =& \frac{\partial {S}_{q,g}}{\partial A_{i j}^k},
\end{align}%
where $ \phi_{i j}^k$ represents a $partial\ linearization$ of the network downstream from $A^{k}_{i j}$ \cite{selvaraju2017grad} and captures the importance of $A_{i j}^k$ for the score $S_{q,g}$.

Like Grad-CAM \cite{selvaraju2017grad}, we calculate a weighted sum over the channel axis and apply ReLU to the combined result to get the salient map $M$. $M_{i j}$ directly correlates with the importance of a particular spatial location $(i, j)$ for whether two images belong to the same identity. The larger $M_{i j}$, the more salient the spatial location $(i, j)$.
\begin{align}
\label{VisMap}
    M_{i j} =& \operatorname{ReLU}(\sum_{k}\phi_{i j}^k A_{i j}^k).
\end{align}%

Naturally, as CG-RAM can get the salient map $M$, it can serve as a visualization method to interpret the ranking result in the test stage. We show some visualization results in Figure \ref{morevis}. Furthermore, as re-ID is a zero-shot learning problem, i.e., the identifications in the evaluation stage do not occur in the training stage, many existing methods \cite{zhou2016learning,selvaraju2017grad,chattopadhay2018grad} designed for the classification task that requires the identifications to find the salient regions cannot be applied in the test stage. In contrast, this problem does not exist in our CG-RAM method.
    
    
    

\subsection{Batch Erasing Salient Method}

In this section, we introduce the Batch Erasing Salient Method; it drops the most salient parts in mini-batch training with a certain probability to make the model discover comprehensive information.

\subsubsection{CG-RAM during Training}
Given a query and a gallery image, CG-RAM can find the salient areas that help the model to identify the two images as the same identity. To find the salient areas, CG-RAM needs the query and galley pairs. However, we cannot directly apply CG-RAM at the training stage. The reasons are as follows: 1) the training set does not have the gallery and query partition; 2) searching the corresponding gallery images in training set for every image is a very time-consuming operation. Specifically, we need to repeat the searching operation every time the model is updated. As a result, the time complexity is $\Theta (N^2E)$, where $N$ represents the training size, and $E$ represents the number of training epoch.

To apply CG-RAM in training, we randomly select $J$ identities and randomly sample $K$ image for each identity in the training set. For each sample $x_{i}$ in the batch, we select the easiest positive sample $x_{p}$ in the $J \times K$ images to represent the galley image of the $x_{i}$. Note that the easiest positive sample, $x_{p}$, has the minimum distance and same identity with the sample $x_{i}$. We can calculate the confidence score $S_{x_{i},x_{p}}$ between the query and galley pair: $x_{i}$ and $x_{p}$.

Different samples, e.g., $x_{i1}$, $x_{i2}$, may have the same easiest positive sample $x_{p}$. As a result, $x_{i1}$ and $x_{i2}$ both have an impact on the salient map of $x_{p}$. To make the pair with higher confidence score $S$ play a more important role in the salient map, we use the $e^{S{x_{i},x_{p}}}$ to represent the weight of pair, and calculate the weighted sum to get the total confidence score $S_{batch}$ in a batch.

\begin{align}
    S_{batch} =& \sum_{i} e^{S{x_{i},x_{p}}} S_{x_{i},x_{p}}.
\end{align}%
Finally, we use Equation \ref{weight} and Equation \ref{VisMap} to get the salient maps. During training, we only need to search in the mini-batch and do one more gradient backpropagation to get the salient maps instead of searching in the whole training set.

\subsubsection{Erasing Method}
Similar to \cite{zhong2017random}, we set an erasing probability $P$ and an erasing ratio $R$. In training, the probability of image $I$ kept unchanged is 1-$P$.

BESM first uses the CG-RAM during the training stage to get the salient map $M$, resize $M$ to the image size, and get $M^{*}$. BESM ranks the pixel $(x,y)$ of the raw images in descending order by $M^{*}(x,y)$ and sets the pixel to 0 if it belongs to the top $R$ value in $M^{*}$. We show the procedure in Algorithm \ref{alg:algorithm}.
\begin{algorithm}[ht]
\caption{Batch Erasing Salient Method}
\label{alg:algorithm}
\textbf{Input}: Erasing Ratio $R$, Erasing Probability $P$, Input Image $I$, Layer to Locate Salient Areas $L$\\
\textbf{Output}: Updated Image $I*$
\begin{algorithmic}[1] 
\STATE Let $P' \leftarrow rand(0,1)$.
\IF {$P' \leq P$}
\STATE Use CG-RAM during training to get the salient map $M$ of the layer $L$.
\STATE Resize the $M$ to the input image size,  and get $M^{*}$

\STATE Rank the pixels $(x,y)$ of $I$ by $M^{*}(x,y)$ in descending order.
\FOR{pixels $(x,y)$ in $I$}
\IF {$M^{*}(x,y)$ belongs to the top $R$ value in $M^{*}$}
\STATE $I(x,y) \leftarrow 0$
\ENDIF
\ENDFOR
\ENDIF
\STATE $I^{*} \leftarrow I$
\STATE \textbf{return} $I^{*}$
\end{algorithmic}
\end{algorithm}

\subsection{P-pooling for Over-Erasing}
It can be difficult for the model to obtain a high-quality and robust feature embedding if the model loses too much important information, especially when GAP is applied in the model. GAP calculates the mean value of each channel. After erasing the salient parts, the remaining areas are not as distinguishable as before. As we mentioned before, GAP cannot effectively extract valuable information from the remaining areas, and GMP is sensitive to noise data.

Based on this phenomenon, we use P-pooling\cite{Wei_2019_CVPR} to alleviate the over-erasing problem, which generalizes GAP and GMP. As a result, P-pooling is robust like GAP and can effectively extract the salient information like GMP.  The result of P-pooling is given by 

\begin{align}
	\label{lehmer}
	   F^k = & P(A^k) =\frac{\sum_{a \in A^k}a^{l}}{\sum_{a \in A^k}a^{l-1}},
\end{align}%
where $F^k$ represents the $k^{th}$ element of the vector $F$, $A^k$ is the $k^{th}$ channel of the feature map $A$, and $l$ is the learnable pooling parameter of P-pooling. Note that GMP and GAP are special cases of P-pooling: P-pooling becomes GMP when $l \rightarrow \infty$, GAP when $l=1$. Since P-pooling is differentiable and can be part of the back-propagation, $l$ can be learned during training.  The corresponding derivative of $l$ is given by
\begin{align}
	\label{lehmer_p}
	  \frac{\partial F^k}{\partial l} = 
		F^k \left[ 
				\frac
					{\sum_{a \in A^k}a^{l}\ln a}
					{\sum_{a \in A^k}a^{l}}
				-
				\frac
					{\sum_{a \in A^k}a^{l-1}\ln a}
					{\sum_{a \in A^k}a^{l-1}}
			\right].
\end{align}%

\section{Experiments}
\subsection{Person re-ID Experiments}

\begin{table*}[tbh]
    \centering
    \caption{Comparison with state-of-the-art methods on Market-1501, DukeMTMC-reID, and MSMT17.} 
    \resizebox{0.75\linewidth}{!}{
        \begin{tabular}{lrrrrrr}
            \toprule
            \multirow{2}{*}{Methods}  & \multicolumn{2}{r}{Market-1501} &  \multicolumn{2}{r}{DukeMTMC-reID}& \multicolumn{2}{c}{MSMT17} \\
            \cmidrule{2-7}
            & rank-1 & mAP  & rank-1 & mAP & rank-1 & mAP  \\
            \midrule
            SVDNet\cite{sun2017svdnet}  & 82.3  & 62.1 &  76.7 & 56.8 &-&- \\
            MGN \cite{wang2018learning} &  \textbf{95.7} & 86.9 &  88.7 & 78.4 &-&- \\
            PCB \cite{sun2018beyond} & 93.8 & 81.6 &  83.3 & 69.2& 68.2&40.4  \\

            CAMA \cite{yang2019towards} & 94.7 & 84.5 &  85.8 & 72.9 &-&- \\
            MHN \cite{chen2019mixed} & 95.1 & 85.0 & 89.1 & 77.2 &-&-\\
            CASN \cite{zheng2019re} & 94.4  & 82.8 & 87.7 & 73.7 &-&-\\ 
            AANet \cite{tay2019aanet} & 93.9 & 83.4 & 87.7 & 74.3 &-&-\\
            DGNet \cite{zheng2019joint} & 94.8 & 86.0 & 86.6 & 74.8 &77.2&52.3\\
            IANet \cite{hou2019interaction} & 94.4 & 83.1 & 87.1 & 73.4 & 75.5 & 46.8\\
            OSNet \cite{zhou2019osnet} & 94.8 & 84.9 & 86.6 & 73.5 & 78.7 & 52.9\\
            \midrule
            Base & 93.9 & 83.2 &  85.0 & 72.1 &75.1 & 47.2 \\
            ES-Net(Ours) & \textbf{95.7} &  \textbf{88.6}  & \textbf{89.2}  & \textbf{78.8} & \textbf{80.9}  & \textbf{57.3} \\
            \bottomrule
        \end{tabular}
    }
\label{table:soa_d_m}
\end{table*}
\subsubsection{Datasets and Evaluation Metrics}
Experiments are conducted on Market-1501 \cite{zheng2015scalable}, DukeMTMC-reID \cite{DBLP:conf/eccv/RistaniSZCT16,zheng2017unlabeled} and MSMT17 \cite{wei2018person}. \textbf{Market1501} contains 12,936 images of 751 identities for training, 3,368 query and 19,732 gallery images of 750 identities for testing. \textbf{DukeMTMC-reID} has 16,522 images of 702 identities for training, 2,228 query and 17,661 gallery images of 702 identities for testing. \textbf{MSMT17} provides 32,621 train images of 1,041 identities, 11,659 query images and 82,161 gallery images of 3,060 identities.

The Rank-1 accuracy and mean Average Precision (mAP) are adopted as evaluation metrics. For each query, its average precision (AP) is computed from its precision-recall curve. The mAP is calculated as the mean value of average precision across all queries. Results with the same identity and the same camera ID as the query image are not counted. For fair comparison, all our results do not use any re-ranking \cite{zhong2017re} or multi-query fusion techniques.

\subsubsection{Implementation Details}
\textbf{Baseline}. We choose the ResNet-50 that is pre-trained on ImageNet as the backbone. The baseline is trained by the cross-entropy loss and uses the techniques discussed in the \protect\cite{luo2019bag}: setting the stride of the last residual block to 1 and adding the BN layer. Our model is implemented on the Pytorch platform.

\textbf{Data Augumentation}. All the images are resized to 384 $\times$ 128, and images are augmented by random horizontal flip and normalization during training. Left-right image flipping is also utilized in the testing stage.

\textbf{Training Strategy}. Each batch is sampled with randomly selected $J$ identities and randomly sampled $K$ image for each identity, and we set $J$ = 16 and $K$ = 4 here. The margin parameter for the triplet loss is 0.3. We adopt Adam as our optimizer with weight decay $5 \times 10^{-4}$. We use the linear warm-up strategy in the first 10 epochs to make the learning rate from 0 to $3.5 \times 10^{-4}$. Then, the learning rate keeps $3.5 \times 10^{-4}$ from 10th epoch to 40th epoch, decays to $3.5 \times 10^{-5}$ at 40th epoch, and decays to $3.5 \times 10^{-6}$ at 70th epoch. The whole training process lasts for 120 epochs. Erasing ratio $R$ is 10\%, and erasing probability $P$ is 0.3. We choose the $res\_conv5$ in the ESB as the layer $L$ to locate the salient parts.

\subsubsection{Comparison with State-of-the-art Methods}
In Table \ref{table:soa_d_m}, we compare our method with current state-of-the-art methods on  Maket1501, DukeMTMC-reID, and MSMT17 datasets.

Our method has yielded overall state-of-the-art performance on all datasets. On Market1501, the rank-1 (95.7 \%) of ES-Net is the same with MGN's rank-1; but mAP of ES-Net (88.6\%) is much better than MGN's (86.9\%) and surpasses all existing methods'. ES-Net outperforms the closest competitor MGN by a large margin of 1.7\% mAP. Besides, MGN mainly benefits from its multi-branch setting, i.e., 8 branches with 11 loss functions. 
On DukeMTMC-reID, ES-Net also achieves state-of-the-art performance (89.2\% rank-1, 78.8\% mAP). On MSMT17, ES-Net obtain the best performance among the compared methods with rank-1 = 80.9\%, mAP = 57.3\%. Compared to the closest competitor OSNet\cite{zhou2019osnet}, we achieve 2.2\%  and 4.4\% improvement in rank-1 accuracy and mAP, respectively.

\subsubsection{Ablation Study}


\begin{table*}[thb]
		\centering
		\caption{The Impact of AIB and ESB on Market-1501, MSMT17, and DukeMTMC-reID. Base: Baseline, AIB: All Information Branch, ESB: Erasing Salient Branch, $R$: the erasing ratio, $P$: the erasing probability.} 
		\resizebox{0.8\linewidth}{!}{
			\begin{tabular}{lrrrrrrr}
				\toprule
				\multirow{2}{*}{Methods}  & \multicolumn{2}{c}{Market-1501} &  \multicolumn{2}{c}{DukeMTMC-reID} & \multicolumn{2}{c}{MSMT17} \\
				\cmidrule{2-7}
				& rank-1 & mAP  & rank-1 & mAP & rank-1 & mAP \\
				\midrule
				Base & 93.9 & 83.2 &  85.0 & 72.1 &75.1 & 47.2 \\
				\midrule
				AIB  & 94.4 & 85.1 & 85.9 & 72.6 & 75.4& 48.0 \\
				ESB\small{($R$=10\%,$P$=0.3)} & 95.2 & 87.1 &  88.6 & 78.0 &79.2 & 54.5 \\
				ESB\small{($R$=20\%,$P$=0.5)} & 94.6 & 86.2 &  86.9 & 77.0 &76.3 & 52.9 \\
				AIB+ESB\small{($R$=10\%,$P$=0.3)} & \textbf{95.7} & 88.0 &  89.1 & 78.1 & \textbf{81.0} & 56.8 \\
				AIB+ESB\small{($R$=20\%,$P$=0.5)} &\textbf{95.7} &  \textbf{88.6}  & \textbf{89.2}  & \textbf{78.8} & 80.9 &  \textbf{57.3} \\
				\bottomrule
			\end{tabular}
		
		}
	\label{table:branch}
\end{table*}

\textbf{The Verification of Motivation} \label{title:prove}

\begin{table*}[thb]
	\centering
	\caption{Ablation studies of the effective components of ESB in ES-Net on Market-1501, MSMT17, and DukeMTMC-reID. Base: Baseline, BESM: Batch Erasing Salient Method, ESB: Erasing Salient Branch.} 
	\resizebox{\linewidth}{!}{
		\begin{tabular}{lrrrrrrr}
			\toprule
			\multirow{2}{*}{Methods}  & \multicolumn{2}{c}{Market-1501} &  \multicolumn{2}{c}{DukeMTMC-reID} & \multicolumn{2}{c}{MSMT17} \\
			\cmidrule{2-7}
			& rank-1 & mAP  & rank-1 & mAP & rank-1 & mAP \\
			\midrule
			Base & 93.9 & 83.2 &  85.0 & 72.1 &75.1 & 47.2 \\
			Base+BESM & 94.3 & 85.1 &  85.7 & 74.3 &76.4 & 49.9 \\
			Base+BESM+P-pooling & 94.6 & 86.4 &  88.2 & 76.6 &78.0 & 52.4 \\
			Base+BESM+P-pooling+Triplet Loss (ESB) & \textbf{95.2} & \textbf{87.1} &  \textbf{88.6} & \textbf{78.0} & \textbf{79.2} & \textbf{54.5} \\
			\bottomrule
		\end{tabular}
	}

\label{table:ablation_esb}
\end{table*}

\begin{figure}[thb]
	\centering
	\includegraphics[width=\linewidth]{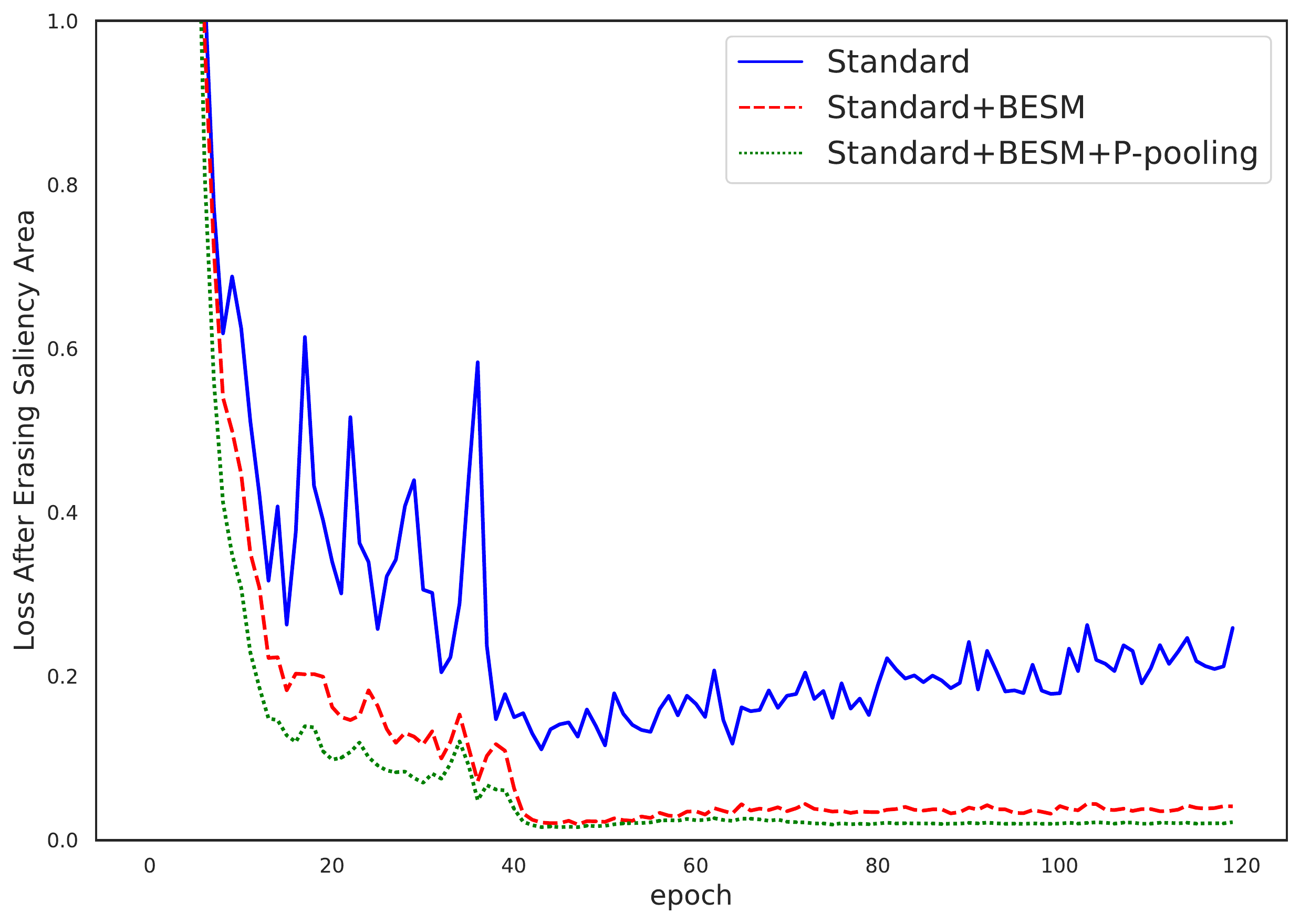}
	\caption{Loss after erasing the salient parts.}
	\label{lossfig}
\end{figure}

We design an experiment to further show that the baseline model pays more and more attention to the salient areas with continuous training. First, we use BESM on the batch image to get the updated image $I^*$, forward the $I^*$ to the model and record the loss. Note that the loss directly reflects the importance of the salient parts to the baseline model, then we use the entire image $I$ of the same batch to train and update the baseline model instead of using  $I^*$. As shown in Figure \ref{lossfig}, the loss of the baseline gradually increases after the 55th epoch, which indicates that the baseline model relies more and more on the salient areas to identify persons. In other words, the baseline model mainly focuses on the salient areas in images. After applying BESM, the magnitude of the loss decreases, but it still has an increasing tendency after the 55th epoch. By contrast, after applying P-pooling, the loss does not increase after the 55th epoch.

\textbf{The Impact of AIB and ESB}
Without the AIB, ESB still performs better than the baseline, as illustrated in Table \ref{table:branch}. The motivation behind the two-branch structure in the ES-Net is that ESB learns more comprehensive features, and AIB prevents losing the discriminative information. After applying AIB, ES-Net can choose a bigger $R$ and $P$ to enable ESB to erase more salient information, discover more information, and achieve better performance. This indicates that the two branches reinforce each other and are both important to the final performance. 

\begin{figure}[thb]
	\centering
	\includegraphics[width=\linewidth]{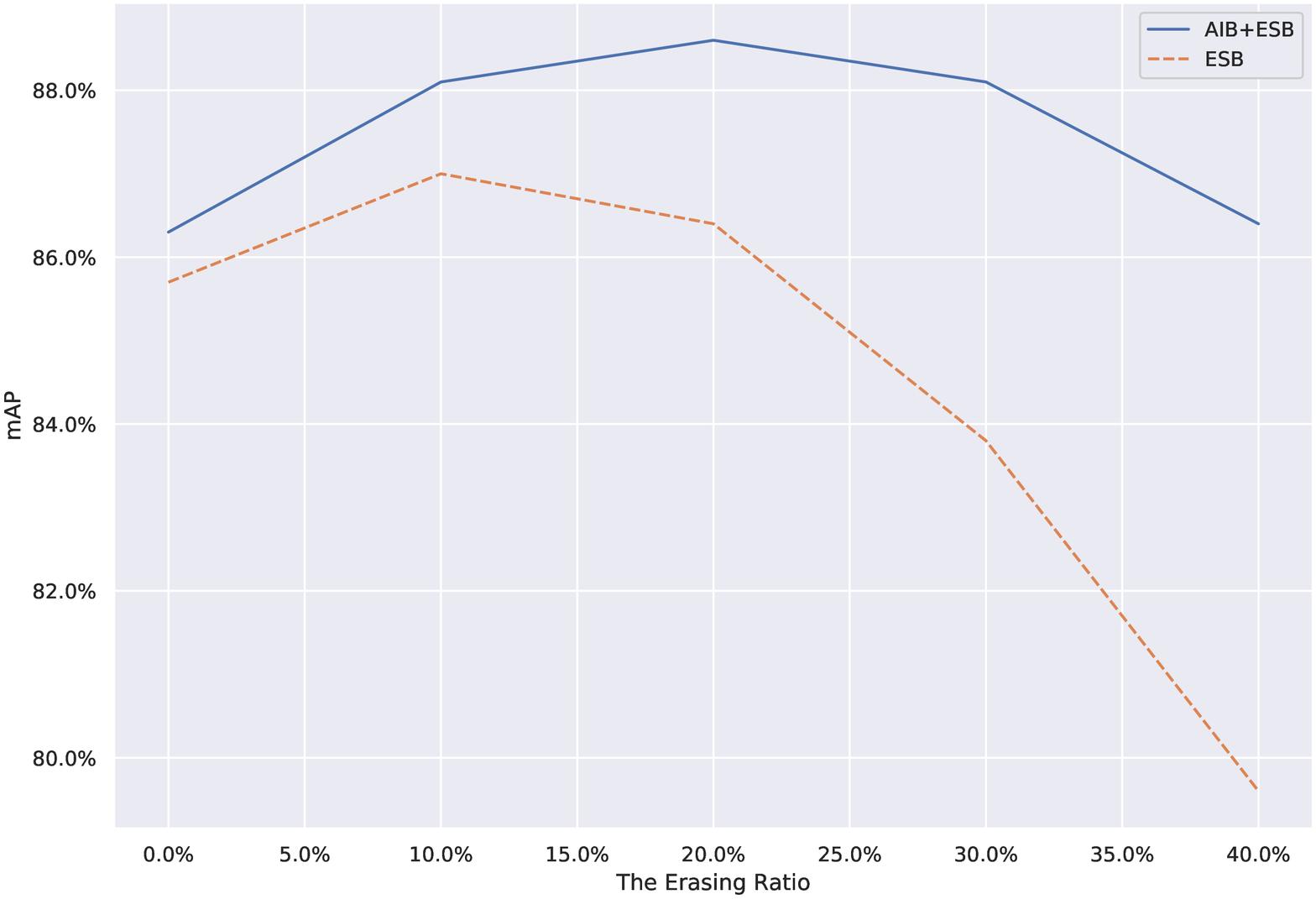}
	\caption{The effects of erasing ratio on mAP for ESB and ESB+AIB.}
	\label{erase_abl}
\end{figure}

To further illustrate the above point, we evaluate different values of $R$ for ESB and AIB+ESB in Fig. \ref{erase_abl}. As $R$ increases from 0 to 40\%, the mAP of ESB and AIB+ESB both increase first and then decrease after erasing too much important information. AIB+ESB gets the best mAP when $R$ is 20\%. However, ESB gets the best mAP when $R$ is 10\%, which means AIB+ESB can use a bigger erasing ratio $R$ than ESB. It is clearly shown that AIB enables our approach to erase more salient parts and discover comprehensive features.


\textbf{The Impact of BESM}
In this section, we investigate the impact of BESM by conducting analytic experiments on three person re-ID datasets. BESM achieves a larger margin over the baseline model, as illustrated in Table \ref{table:ablation_esb}. Specifically, BESM improves the 1.9\% mAP and 0.4\% rank-1 on Market-1501, the 2.2\% mAP and 0.7\% rank-1 on DukeMTMC-reID and 2.7\% mAP and 1.3\% rank-1 on MSMT17. Meanwhile, compared to baseline, BESM has a much smaller error with continuous training, as shown in Figure \ref{lossfig}. These results validate the powerful capability of the proposed BESM to force the ESB to discover comprehensive features by erasing the salient parts. 

Besides, in Fig. \ref{fig:viscmp}, the CG-RAM shows the associated visual cues between query and gallery images. We can observe that for the same input query image, the features learned by Baseline+BESM are indeed more diverse than baseline's.  All these indicate that the BESM forces the model to learn more and boosts the performance.

\begin{table*}[thb]
	\centering
	\caption{Comparison with other pooling methods on Market-1501, DukeMTMC-reID, and MSMT17. GAP: global average pooling, GMP: global max pooling.
	}  
	\resizebox{0.85\linewidth}{!}{
		\begin{tabular}{lrrrrrrr}
			\toprule
			\multirow{2}{*}{Methods}  & \multicolumn{2}{c}{Market-1501} &  \multicolumn{2}{c}{DukeMTMC-reID} & \multicolumn{2}{c}{MSMT17} \\
			\cmidrule{2-7}
			& rank-1 & mAP  & rank-1 & mAP & rank-1 & mAP \\
			\midrule
			Base+BESM+GAP & 94.3 & 85.1 &  85.7 & 74.3 &76.4 & 49.9 \\
			Base+BESM+GMP &  94.2& 84.4 & 87.2  & 75.3 &75.0 & 48.5 \\
			\midrule
			Base+BESM+P-pooling & 94.6 & 86.4 &  88.2 & 76.6 &78.0 & 52.4 \\
			ES-Net(Ours) & \textbf{95.7} &  \textbf{88.6}  & \textbf{88.5}  & \textbf{78.7} & \textbf{80.5} &  \textbf{57.3} \\
			\bottomrule
		\end{tabular}
	
	}

\label{table:cmp_pooling}
\end{table*}

\textbf{The Impact of P-pooling}
P-pooling is a vital part of the ES-Net since ESB may not efficiently extract valuable information if we erase too many salient areas. In Table \ref{table:ablation_esb}, we can see that the P-pooling significantly improves performance. In particular, P-pooling improves the 1.3\% mAP and 0.3\% rank-1 on Market-1501, the 2.3\% mAP and 2.5\% rank-1 on DukeMTMC-reID and 2.5\% mAP and 1.6\% rank-1 on MSMT17.

Moreover, as shown in Figure \ref{lossfig}, the loss of BESM still slowly increases after the 55th epoch, which is caused by the over-erasing problem. After applying P-pooling, the loss does not increase after the 55th epoch, which demonstrates that P-pooling mitigates the over-erasing problem.

\textbf{Comparison with Other Erasing Methods}
We compare our method with other two erase work. One is called Random Erasing \cite{zhong2017random}, and it erases part of images randomly. The other is Feature Dropping Branch \cite{dai2019batch}, which randomly erases part of feature maps in the training batch.  As can be seen in Table \ref{table:cmp_other_re}, our BESM is much better than Random Erasing and Feature Dropping Branch. Specifically, on Market-1501,
BESM improves the rank-1 accuracy and mAP by 0.6 points and 1.9 points, respectively. On MSMT17, the rank-1 accuracy and mAP are improved
2.3 points and 1.8 points, respectively.

\begin{table}[thb]
	\centering
	\caption{Comparison with other erase methods on Market-1501, and MSMT17. Base: Baseline, BESM: Batch Erasing Salient Method. Grad-CAM Erasing means that we use Grad-CAM to get and erase the salient parts. Compared with our proposed CG-RAM, Grad-CAM is designed for the classification task.}  
	\resizebox{\linewidth}{!}{
		\begin{tabular}{lrrrrr}
			\toprule
			\multirow{2}{*}{Methods}  & \multicolumn{2}{c}{Market-1501}  & \multicolumn{2}{c}{MSMT17} \\
			\cmidrule{2-5}
			& rank-1 & mAP  & rank-1 & mAP  \\
			\midrule
			Feature Dropping Branch \cite{dai2019batch} & 93.6 & 83.3  &- & - \\
			Base+Random Erasing \cite{zhong2017random} & 92.5 & 82.9 &  72.8 & 48.1  \\
			Base+Grad-CAM Erasing & 94.1 & 84.7 &  74.6 & 49.4  \\
			Base+Feature Dropping Branch\cite{dai2019batch}  & 93.7 & 83.2  &73.1 &47.0 \\
			\midrule
			Base+BESM & 94.3 & 85.1 &  75.4 & 49.9  \\
			ES-Net(Ours) &\textbf{95.7} &  \textbf{88.6}  & \textbf{80.5} &  \textbf{57.3} \\
			\bottomrule
		\end{tabular}
	
	}
\label{table:cmp_other_re}
\end{table}

\textbf{BESM vs Grad-CAM Erasing}
To further demonstrate the effectiveness of our method, we also implement Grad-CAM Erasing which uses Grad-CAM to locate the salient map by cross-entropy loss and erases the salient parts during training. The salient visual information for the re-ID task is also likely to be the salient parts for the classification task, and the images of re-ID are low-resolution. As a result, the Grad-CAM and our proposed CG-RAM may get similar salient maps and erase the same salient parts. However, our proposed CG-RAM better reflects the retrieval process of re-ID due to the supervision of the confidence score $S$, which makes BESM get better results than Grad-CAM erasing. In particular, our BESM outperforms Grad-CAM Erasing 0.4\% mAP on Market-1501 and 0.5\% mAP on MSMT-17, as shown in in Table \ref{table:cmp_other_re}. Overall, ES-Net outperforms Grad-CAM Erasing significantly.

\begin{figure}[t]
	\centering
	\includegraphics[width=\linewidth]{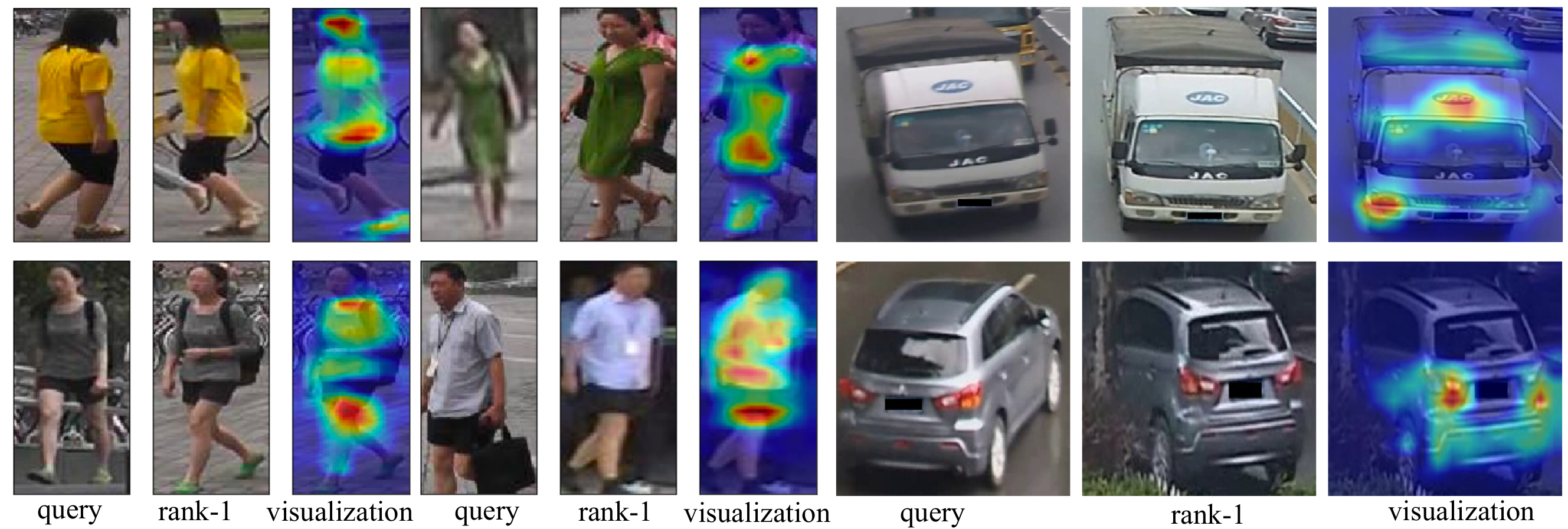}
	\caption{Visualization results on person re-ID and vehicle re-ID. The "Visualization" is the visualization results of the salient areas in the query images, which are found by proposed CG-RAM. Note that the visualized salient areas found by CG-RAM show human-interpretable visual explanations for the ranking results.}
	\label{morevis}
\end{figure}

Additionally, Grad-CAM cannot be used to visualize and interpret the ranking results during testing, while CG-RAM is still applicable. We show some examples of the visualization results generated by CG-RAM in Fig. \ref{morevis}, which indicates the associated visual features between query and gallery images. In particular, yellow bags in the third pair of the first row and green clothes in the third pair of the second row are the salient features according to the CG-RAM. It is obvious that the visualized salient areas found by CG-RAM show human-interpretable visual explanations for the ranking results.

\begin{table*}[htb]
    \centering
    \caption{The Impact of applying p-Pooling in AIB on Market-1501, DukeMTMC-reID, and MSMT17. ES-Net(origin): Network structure proposed in the main text that applies GAP in AIB. ES-Net(apply P-pooling in AIB and ESB): Network structure that applies P-pooling in AIB and ESB} 
    \resizebox{\linewidth}{!}{
        \begin{tabular}{lrrrrrr}
            \toprule
            \multirow{2}{*}{Methods}  & \multicolumn{2}{r}{Market-1501} &  \multicolumn{2}{r}{DukeMTMC-reID}& \multicolumn{2}{c}{MSMT17} \\
            \cmidrule{2-7}
            & rank-1 & mAP  & rank-1 & mAP & rank-1 & mAP  \\
            \midrule

            ES-Net(origin) & 95.7 &  \textbf{88.6}  & \textbf{89.2}  & 78.8 & 80.9  & \textbf{57.3} \\
             ES-Net(apply P-pooling in AIB and ESB) & \textbf{95.8} &  88.3  & 88.9  & \textbf{79.3} & \textbf{81.1}  & \textbf{57.3} \\
            \bottomrule
        \end{tabular}
    }
\label{table:p-pooling}
\end{table*}

\begin{table}[thb] 
	\centering
	\caption{Applying Stronger Backbone, i.e. AGW\cite{ye2020deep}, in ES-Net on Market-1501, DukeMTMC-reID and MSMT17. ES-Net$^{\star}$ means that applying AGW as the backbone in the ES-Net instead of resnet-50.
	}  
	\resizebox{\linewidth}{!}{
		\begin{tabular}{lrrrrrrr}
			\toprule
			\multirow{2}{*}{Methods}  & \multicolumn{2}{c}{Market-1501} &  \multicolumn{2}{c}{DukeMTMC-reID} & \multicolumn{2}{c}{MSMT17} \\
			\cmidrule{2-7}
			& rank-1 & mAP  & rank-1 & mAP & rank-1 & mAP \\
			\midrule
			AGW & 95.1 & 87.8 &  89.0 & 79.6 &68.3 & 49.3 \\
			ES-Net$^{\star}$ & \textbf{95.3} &  \textbf{89.1}  & \textbf{89.7}  & \textbf{80.2} & \textbf{77.9} &  \textbf{55.6} \\
			\bottomrule
		\end{tabular}
	
	}

\label{table:agw}
\end{table}

\textbf{Comparison with Other Global Pooling}
We also compare P-pooling with other global pooling methods in Table \ref{table:cmp_pooling}. Our proposed P-pooling achieves the best performance. 
Specifically, after apply P-pooling, our method achieves 86.4\% mAP, 94.6\% rank-1 on Market1501, 76.6 \%mAP, 88.2\%rank-1 on DukeMTMC-reID and 52.4\% mAP, 78.0\% rank-1 on MSMT17. P-pooling outperforms GAP and GMP by a large margin (1.3\% mAP, 0.3\% rank-1 on Market1501, 1.3\%mAP, 0.5\%rank-1 on DukeMTMC-reID and 2.5\% mAP, 1.6\% rank-1 on MSMT17).

\textbf{Apply P-pooling in AIB}
We can use LMP in AIB. However, AIB uses the entire pictures, which means the over-erasing problem is not in AIB. So GAP can already achieve good enough results, and the results of GAP and P-pooling are also almost the same. 

To further prove our point, we conducted a specific experiment about applying P-pooling in AIB of ES-Net.  As shown in TABLE \ref{table:p-pooling}, the results of applying P-pooling in AIB are very close to the results of using GAP in AIB.

\begin{figure}[t]
    \centering
    \includegraphics[width=\linewidth]{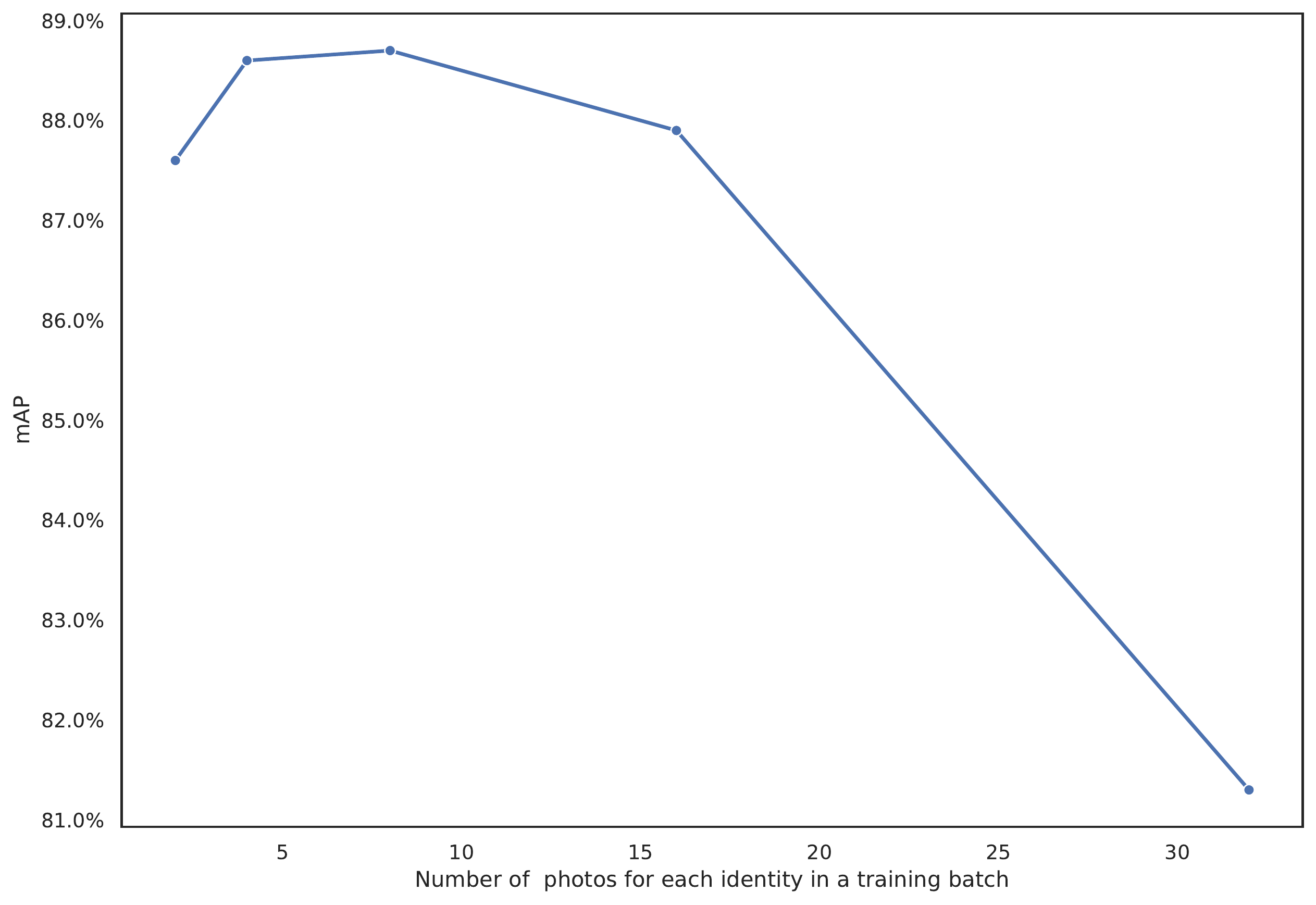}
    \caption{The Impact of different J and K on Market-1501. J: Number of different identities in a training batch, K: Number of photos for each identity in a training batch. Note that We set the batch size equal to 64 unchanged in this experiment.
    }
    \label{fig:J_k}
    \centering
\end{figure}

\textbf{Impact of Hyper-parameters $J$ and $K$}
Fig. \ref{fig:J_k} studies the impact of  Hyper-parameters $J$ and $K$ on the performance of the ES-Net. Here, $J$ is the number of different identities in a training batch, $K$ is the number of photos for each identity in a training batch. Note that We set the batch size equal to 64 unchanged in this experiment. $J=16$ and $K=4$ is the common setting in re-ID, which is the setting for ES-Net in this paper.

\textbf{Stronger Backbone AGW\cite{ye2020deep}}
To further illustrate our method's effectiveness, we use a stronger backbone, AGW\cite{ye2020deep}, instead of ResNet-50. The results are shown in TABLE \ref{table:agw}. As we can see, ES-Net$^{\star}$ can still improve accuracy. ES-Net$^{\star}$ means that applying AGW as the backbone in the ES-Net instead of resnet-50. In particular, ES-Net$^{\star}$ improves the 1.3\% mAP and 0.2\% rank-1 on Market-1501, the 0.6\% mAP and 0.7\% rank-1 on DukeMTMC-reID.

\subsection{Vehicle Re-identification Experiments}

\subsubsection{Datasets and Settings}

The ES-Net can be applied directly to vehicle re-identification problems. We conduct extensive experiments on  VehicleID\cite{liu2016deep} and Veri-776\cite{liu2016a}. \textbf{VehicleID} is a large-scale vehicle dataset. The training set contains 10,178 images of 13,134 vehicles, and the test set includes 111,585 images of 13,113 other vehicles. The test set is divided into three subsets with different sizes (i.e., Small, Medium, Large). \textbf{Veri-776} consists of more than 50,000 images of 776 vehicles captured by 20 surveillance cameras. The training set contains 37,778 images of 576 vehicles, and the test set includes 11,579 images of 200 vehicles. On Veri-776, we follow the person re-ID's protocol to evaluate results, which means that query images and the correct gallery images must be captured from different cameras.

The input images are resized to $256 \times 256$. The input images are augmented by random horizontal flip and normalization during training. Left-right image flipping is also utilized in the testing stage. The whole training process lasts for 60 epochs, and the learning rate is decreased by a factor of 0.1 after the 20th and 40th epoch. Erasing ratio $R$ of ES-Net is 20\%, and erasing probability $P$ of ES-Net is 0.5. We choose the $res\_conv5$ in the ESB as the layer $L$ to locate the salient parts. The other settings of vehicle re-identification are the same as person re-ID's.

\begin{table*}[thb]
    \vspace{0.13cm}
	\centering
	\caption{Comparison with state-of-the-art methods on VehicleID. * denotes the paper uses the self-designed network structure. ESB: Base+BESM+P-pooling+Triplet Los. ES-Net: ESB+AIB} 
	\vspace{0.13cm}
		\resizebox{0.9\linewidth}{!}{
			\begin{tabular}{lrrrrrrr}
				\toprule
				\multirow{2}{*}{Methods} & \multirow{2}{*}{Backbones} & \multicolumn{2}{c}{VehicleID (Small)}& \multicolumn{2}{c}{VehicleID (Medium)}& \multicolumn{2}{c}{VehicleID (Large)} \\
				\cmidrule{3-8}
				&& rank-1 & rank-5 & rank-1 & rank-5& rank-1 & rank-5  \\
				\midrule
				CLVR \cite{Kanaci2017KANACIZG} & Inception-V3 &  62.0& 76.0 & 56.1 & 71.8 & 50.6 & 68.0 \\
				C2F \cite{Guo2018LearningCS} &GoogLeNet &  61.1& 81.7 &  56.2& 76.2 &  51.4& 72.2  \\
				Ram \cite{ramVe} &* &  75.2& 91.5 & 72.3 & 87.0 & 76.7 & 84.5 \\
				NuFACT \cite{8036238} &* &  48.9& 69.5 & 43.6 & 65.3 & 38.6 & 60.7 \\
				
				AAVER \cite{Khorramshahi_2019_ICCV} &ResNet-50&  72.5 & 93.2 &66.9&89.4&60.2&84.9 \\
				SAN \cite{qian2020stripe}  & ResNet-50& 79.7 & 94.3 & 78.4 & 91.3 & 75.6 & 88.3\\ 
Part \cite{he2019part}  & ResNet-50& 78.4 & 92.3 & 75.0 & 88.3 & 74.2 & 86.4 \\
	VAMI \cite{zhouy2018viewpoint} & * & 63.1 & 83.3 & 52.9 & 75.1 & 47.3 & 70.3\\
				\midrule
				Base   & ResNet-50& 79.1 & 93.3 &76.8&88.6& 73.4 & 85.7 \\
				ESB(Ours) & ResNet-50& 80.5  & 94.7 & 77.6  & 90.3& 74.4 & 87.4 \\
				ES-Net(Ours) & ResNet-50& \textbf{83.6}  & \textbf{96.9} & \textbf{79.9}  & \textbf{93.5}&\textbf{76.9}&\textbf{90.7} \\
				\bottomrule
			\end{tabular}
	}
\label{table:soa_d_m_v}
\vspace{0.13cm}
\end{table*}

\begin{table}[t]
	\centering
	\caption{Comparison with state-of-the-art methods on Veri-776. * denotes the paper uses the self-designed network structure. ESB: Base+BESM+P-pooling+Triplet Los. ES-Net: ESB+AIB.} 
		\resizebox{\linewidth}{!}{
			\begin{tabular}{lrrrr}
				\toprule
				\multirow{2}{*}{Methods}  &\multirow{2}{*}{Backbones}  & \multicolumn{3}{c}{VeRi-776}  \\
				\cmidrule{3-5}
				&& rank-1 & rank-5 & mAP   \\
				\midrule
				Ram \cite{ramVe} &* &  88.6& 94.0 &61.5 \\
				NuFACT \cite{8036238} &* &  81.6& 95.1 &53.4 \\
				
				AAVER \cite{Khorramshahi_2019_ICCV} &ResNet-50 &  88.7& 94.1 & 58.5  \\
				SAN \cite{qian2020stripe}  &ResNet-50 & 93.3&  97.1& 72.5 \\ 
Part \cite{he2019part} &ResNet-50 & 94.3& \textbf{98.7}& 74.3 \\
	PAMTRI \cite{tang2019pamtri}& DenseNet121& 92.9 & 97.0 & 71.9 \\
	VAMI \cite{zhouy2018viewpoint}  & * &50.1 & 90.8& 77.0 \\
				\midrule
				Base   &ResNet-50& 95.0 & 97.6& 74.4  \\
				ESB(Ours)   &ResNet-50& 96.4 & 98.2& 79.2  \\
				ES-Net(Ours) &ResNet-50 & \textbf{97.0} & 98.5&  \textbf{81.9}  \\
				\bottomrule
			\end{tabular}
	}
\label{table:soa_d_m_veri}
\end{table}

\subsubsection{Comparison with State-of-the-Art}

Table \ref{table:soa_d_m_v} shows that our method achieves the best scores on all datasets on VehicleID. In particular, ES-Net get 83.6\% rank-1, 96.9\% rank-5 on VehicleID (Small), 79.9 rank-1, 93.5\% rank-5 onVehicleID (Medium) and 76.9\% rank-1, 90.7\% rank-5 on VehicleID (Large). 

On VeRi-776, ES-Net also achieves state-of-the-art performance. As shown in table \ref{table:soa_d_m_veri}, ES-Net obtains the state-of-the-art performance among the compared methods with rank-1 = 97.0\%, rank-5 = 98.5\%, mAP = 81.9\%. Compared to the closest competitor, we achieve 2.7\%, 4.9\% improvement in rank-1 accuracy and mAP, respectively. On VeRi-776, the rank-5 (98.5\%) of ES-Net is smaller than Part's rank-5 (98.7\%); but mAP of ES-Net (81.9\%) and rank-1 of ES-Net (97.0\%) is much better than Part's (94.3\% rank-1, 74.3\% mAP)

\subsubsection{Ablation Study}

To further demonstrate the effectiveness of our method on Vehicle Re-identification,  we conduct the experiments of the effective components of ESB on the Veri dataset. As shown in TABLE. \ref{table:ablation_esb_veri},  BESM improves the mAP by 2.1 points and P-pooling improves the mAP by 0.5 points.

\begin{table}[t]
	\centering
	\caption{Ablation studies of the effective components of ESB  on Veri. Base: Baseline, BESM: Batch Erasing Salient Method, ESB: Erasing Salient Branch.} 
\resizebox{\linewidth}{!}{
			\begin{tabular}{lrrr}
				\toprule
				\multirow{2}{*}{Methods}  &   \multicolumn{3}{c}{VeRi-776}  \\
				\cmidrule{2-4}
				& rank-1 & rank-5 & mAP   \\
				\midrule
				
				Base   & 95.0 & 97.6& 74.4  \\
				Base+BESM & 95.9 & 98.3& 76.5  \\
				Base+BESM+P-pooling & 96.0 & 98.1& 77.0  \\
				Base+BESM+P-pooling+Triplet Loss & 96.4 & 98.2& 79.2  \\
				\bottomrule
			\end{tabular}
	}

\label{table:ablation_esb_veri}
\vspace{-0.45cm}
\end{table}

\section{Conclusion}
In this paper, we propose ES-Net to discover comprehensive features by erasing the salient parts. To find the salient parts, we introduce a novel method CG-RAM that utilizes the gradients of the confident score with respect to the features. CG-RAM also can be used to visualize and interpret the ranking results during testing. Based on CG-RAM, we propose BESM to erase the salient parts efficiently in mini-batch during training. Meanwhile, to mitigate the over-erasing problem, we use P-pooling that generalizes GMP and GAP. Through considerable experiments on three Person re-ID datasets and two Vehicle re-ID datasets, we show that ES-Net captures comprehensive features and achieves state-of-the-art performance. In the future, we will extend the idea to other common tasks.



\section*{Acknowledgment}


This work was supported in part by The National Key Research and Development Program of China (Grant Nos: 2018AAA0101400), in part by The National Nature Science Foundation of China (Grant Nos: 62036009, U1909203, 61936006), in part by the Alibaba-Zhejiang University Joint Institute of Frontier Technologies


%





\ifCLASSOPTIONcaptionsoff
  \newpage
\fi





\bibliographystyle{IEEEtran}
\bibliography{IEEEabrv,ES-Net}
%

\begin{IEEEbiography}[{\includegraphics[width=1in,height=1.25in,clip,keepaspectratio]{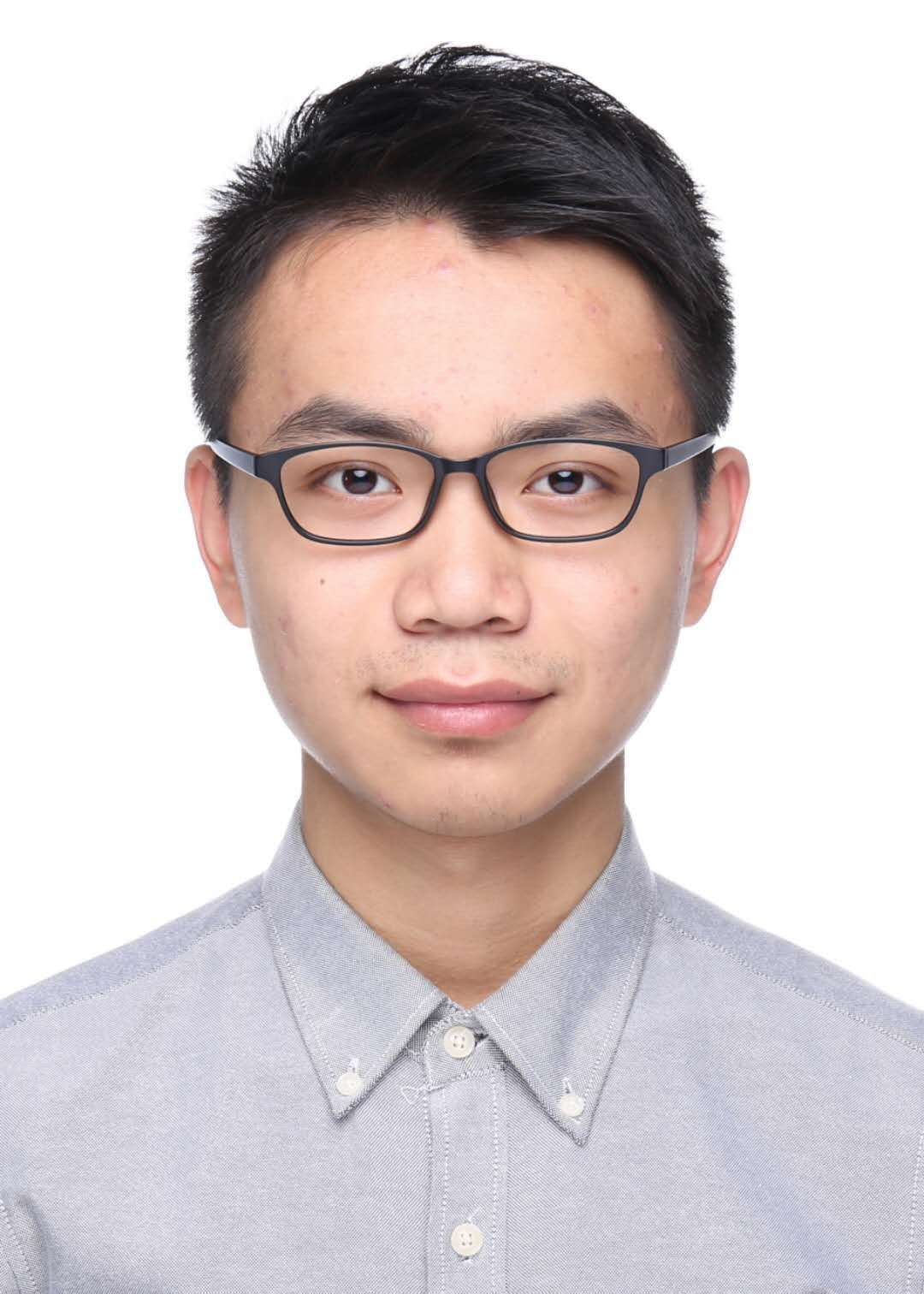}}]{Dong shen}
received the B.S. degree from in Computer Science from Zhejiang University , China, in 2018. He is now a graduate student at Zhejiang University and his research interests include computer version and machine learning.
\end{IEEEbiography}
\begin{IEEEbiography}[{\includegraphics[width=1in,height=1.25in,clip,keepaspectratio]{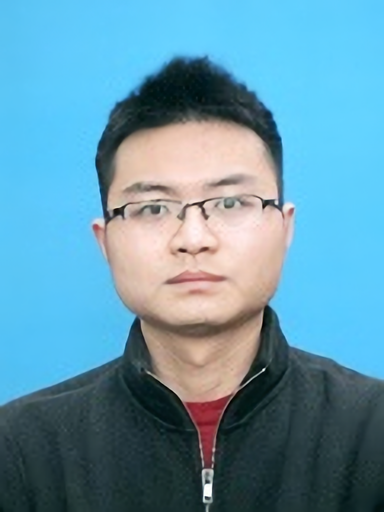}}]{Shuai Zhao} received the Master degree in Computer Science from Zhejiang University. His research interests include computer vision, machine learning and data mining.
\end{IEEEbiography}
\begin{IEEEbiography}[{\includegraphics[width=1in,height=1.25in,clip,keepaspectratio]{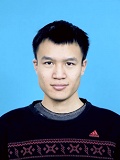}}]{Jingming Hu} received the Master degree in Computer Science from Zhejiang University. His research interests include computer vision, machine learning and data mining.
\end{IEEEbiography}
\begin{IEEEbiography}[{\includegraphics[width=1in,height=1.25in,clip,keepaspectratio]{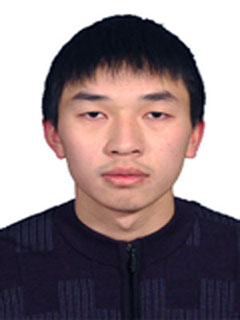}}]{Hao Feng} received the B.S. degree from in Computer Science from Zhejiang University , China, in 2018. He is now a graduate student at Zhejiang University and his research interests include computer version and machine learning.
\end{IEEEbiography}
\begin{IEEEbiography}[{\includegraphics[width=1in,height=1.25in,clip,keepaspectratio]{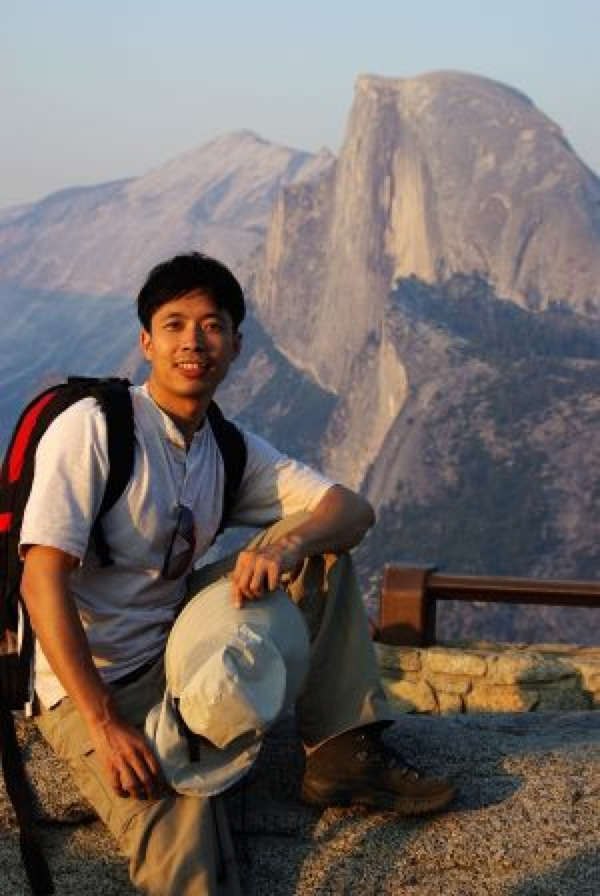}}]{Deng Cai}
is currently a full professor in the College of Computer Science at Zhejiang University, China. He received the PhD degree from University of Illinois at Urbana Champaign. His research interests include machine learning, computer vision, data mining and information retrieval.
\end{IEEEbiography}
\begin{IEEEbiography}[{\includegraphics[width=1in,height=1.25in,clip,keepaspectratio]{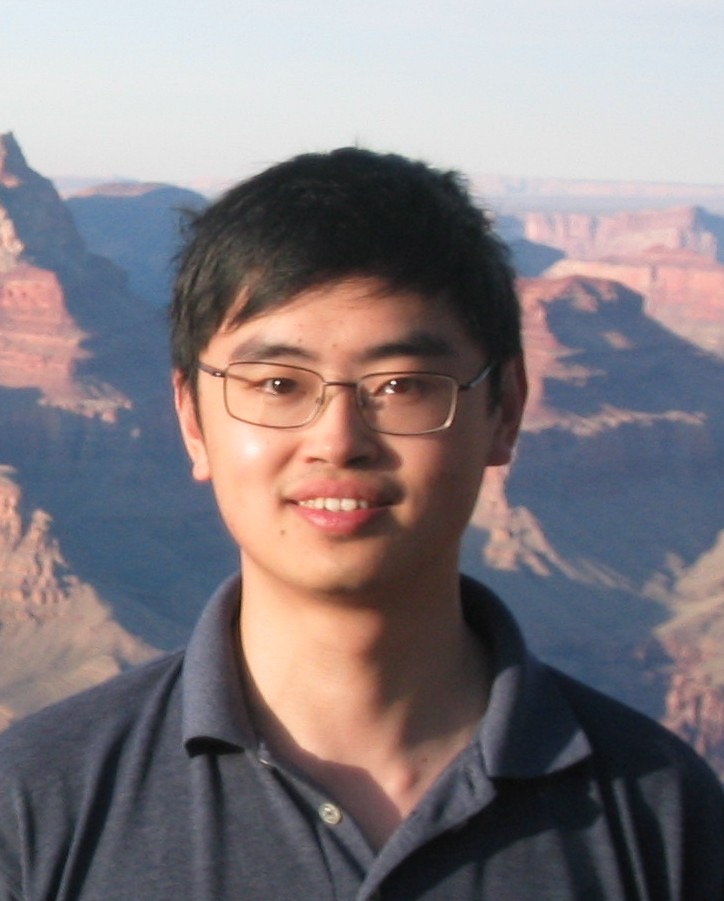}}]{Xiaofei He}
is currently a full professor in the College of Computer Science at Zhejiang University, China. He received the BS degree from Zhejiang University, China, in 2000 and the PhD degree from University of Chicago in 2005, both in computer science. After my PhD, He joined Yahoo Research Labs as a research scientist. He joined Zhejiang University in 2007.
\end{IEEEbiography}




\vfill


\end{document}